%% file: main.tex
\documentclass[10pt, a4paper]{article}

\usepackage{lrec2026} 

\usepackage{booktabs}
\usepackage{listings}
\usepackage{xcolor}
\usepackage{subcaption}
\usepackage{multirow}
\usepackage{rotating}
\usepackage{subcaption}
\usepackage{enumerate}

\colorlet{punct}{red!60!black}
\definecolor{background}{HTML}{EEEEEE}
\definecolor{delim}{RGB}{20,105,176}

\lstdefinelanguage{json}{
	basicstyle=\small\fontfamily{ccr}\selectfont,
	showstringspaces=false,
	breaklines=true,
	frame=lines,
	backgroundcolor=\color{background},
	literate=
	*{:}{{{\color{punct}{:}}}}{1}
	{,}{{{\color{punct}{,}}}}{1}
	{\{}{{{\color{delim}{\{}}}}{1}
	{\}}{{{\color{delim}{\}}}}}{1}
	{[}{{{\color{delim}{[}}}}{1}
	{]}{{{\color{delim}{]}}}}{1}
	{ö}{{\"o}}{1}
	{ä}{{\"a}}{1}
	{ü}{{\"u}}{1}
	{Ö}{{\"O}}{1}
	{Ä}{{\"A}}{1}
	{Ü}{{\"U}}{1}
	{_}{{\_}}{1}
	{ß}{{\ss}}{1},
}

\title{Retrieving Floods without Floodlights: Topic Models as Binary Classifiers for Extreme Climate Events in German News}

\name{Brielen Madureira$^{1,2}$, Mariana Madruga de Brito$^2$, Andreas Niekler$^{1,3}$} 

\address{\\$^1$LeipzigLab - Climate Discourse, Leipzig University, Germany \\
         $^2$Helmholtz Centre for Environmental Research - UFZ, Germany\\
         $^3$ Computational Humanities, Leipzig University, Germany \\ \\
         \texttt{brielen.madureira@uni-leipzig.de}\\
         \texttt{mariana.brito@ufz.de}\\
         \texttt{aniekler@informatik.uni-leipzig.de}\\}

\abstract{
In studies of media coverage of extreme climate events, NLP methods have become indispensable for identifying relevant texts in large news databases. Still, enough annotated data to train accurate deep learning-based classifiers from scratch is often not available. Topic Models have the advantage of being both unsupervised and interpretable, but are typically used only for exploratory analysis or data characterisation. In this study, we investigate how to employ Topic Models as binary classifiers for refining the retrieval of relevant news about seven types of extreme climate events in the German media. Our method relies on the \textit{posterior} distributions estimated by Topic Models to select relevant documents, without modifying their training procedure. Using an annotated sample to guide the evaluation, we show that the probabilities assigned to keywords used to query news databases can also be informative for selecting relevant topics and improve sample precision. We compare our results to a fine-tuned text embedding classifier and an open-weight LLM, discussing observed trade-offs, e.g.~the LLM's lowest precision. Moreover, we show that results are hazard-dependent, which speaks against considering climate events as a single category in NLP tasks.
 \\ \newline }

\begin{document}

\maketitleabstract

\section{Introduction}
\label{sec:intro}
\input{contents/intro}

\section{Related Literature}
\label{sec:lit}
\input{contents/literature}

\section{Methods}
\label{sec:methods}
\input{contents/method}

\section{Data}
\label{sec:data}
\input{contents/data}

\section{Experiments}
\label{sec:experiments}
\input{contents/experiments}

\section{Results}
\label{sec:results}
\input{contents/results}

\section{Analysis}
\label{sec:analysis}
\input{contents/analysis}

\section{General Discussion}
\label{sec:discussion}
\input{contents/discussion}

\section{Conclusion}
\label{sec:conclusions}
\input{contents/conclusions}

\section*{Limitations}
\label{sec:limitations}
\input{contents/limitations}

\section*{Acknowledgements}
We thank Marc Keuschnigg for his contribution in conceptualising the research project that motivates this paper, as well as Maike Reichel and Julius Hehenkamp for their help in annotating the data. We also thank the anonymous reviewers for their valuable feedback.

\section{Bibliographical References}\label{sec:reference}

\bibliographystyle{lrec2026-natbib}
\bibliography{bibs/custom,bibs/anthology-1}

\appendix

\section{Appendix}
\label{sec:appendix}
\input{contents/appendix}

\end{document}

%% file: contents/intro.tex
Assume we are gathering news about floods events to study collective attention in the media. Simply querying a news database to retrieve documents containing the string \textit{flood} would not only match news reporting on actual floods, but also many false positives. Consider this (obviously constructed) example: ``\textit{Soccer fans experienced a flood of emotions witnessing floodlights being turned on as players flooded the field: the game could finally begin after the risk of a flash flood was ruled out.}'' This illustrates a central challenge in information retrieval: the term \textit{flood} can have metaphorical senses, be part of compound nouns unrelated to climate or refer to a merely hypothetical hazard. Thus, despite the repeated presence of the term \textit{flood}, this text is rendered unrelated to actual flood events. 

Pitfalls like that can emerge at the intersection of environmental and social sciences, such as in text-based climate impact and adaptation research. This field often relies on NLP methods to process texts about climate events and their consequences \citep[\textit{inter alia}]{henrique_lima_alencar_flash_2024,nunes_carvalho_unveiling_2024}. In this context, dictionary-based retrieval is a typical procedure: large databases are queried using a curated list of hazard-related keywords to find potentially relevant documents about  e.g.~floods, droughts or wildfires (e.g.~\citealp{sodoge_automatized_2023,egusphere-2025-4891}). But as we just saw above, the mere presence of a keyword in a document does not guarantee its relevance. If term presence or frequency are directly used as predictors in quantitative assessments, research validity is impaired.

\begin{figure}
	\includegraphics[trim={0cm 3.5cm 12.5cm 0},clip,width=\linewidth]{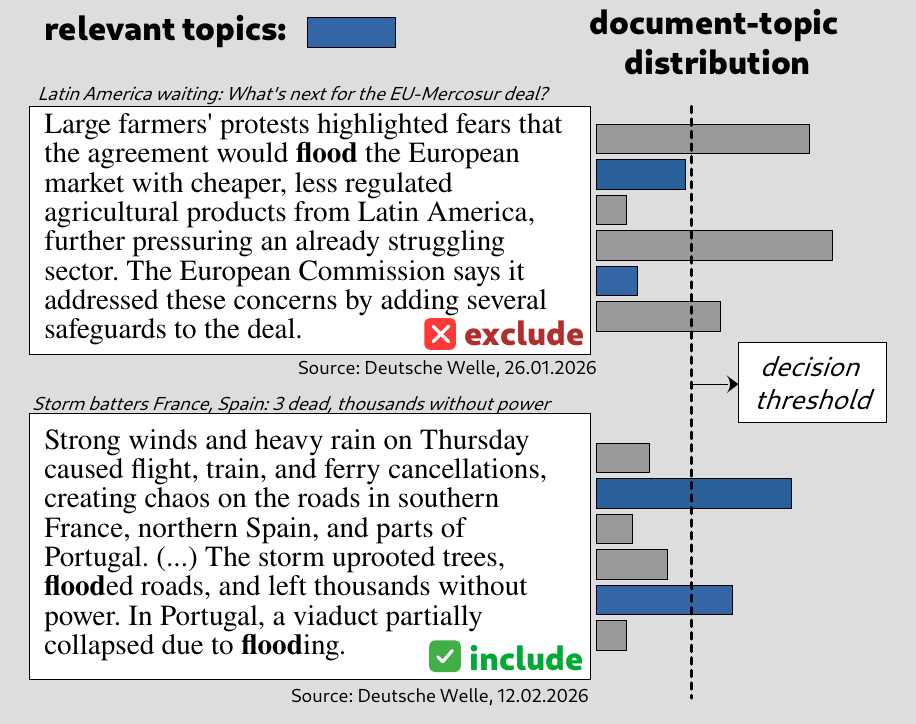}
	\caption{Relevant news articles can be identified based on the saliency of relevant topics in their representation estimated by a topic model.}
	\label{fig:intro-overview}
\end{figure}

Although keyword-based matching serves as a reasonable prefiltering step for creating an initial collection of documents with good recall, retrieval results must be further refined using other classification methods to detect true and false positives and improve their precision. This is an open problem recently discussed by \citet{grasso-etal-2024-nytac}. 

In this paper, we investigate the possibility of yielding a binary classification model for identifying relevant documents using the probabilities estimated by unsupervised vanilla Topic Models (TM), as the overview in Figure \ref{fig:intro-overview}. We assume a situation with a small amount of annotated data that is not enough for training deep learning-based models from scratch but still informative for evaluation. 

Our main contributions in this paper are: 

\begin{enumerate}[i]
	\item a data analysis of news articles in German annotated with seven types of extreme climate events;
	
	\item the usage of TMs for relevance classification without any needed modification on the training regime and no direct human effort in topic interpretation; and
	
	\item evidence that TMs are, for some hazards, on par with deep learning alternatives, with the advantage of interpretability and a tendency to higher precision.
\end{enumerate}

%% file: contents/literature.tex
\paragraph{Retrieval of environment-related documents} Document retrieval is an ubiquitous step in creating corpora for socio-environmental research. To name a few recent large-scale approaches, \citet{leippold2020climatext} implemented a graph-based heuristic on Wikipedia metadata of entries on climate topics, \citet{Kong07022026} relied on climate-related keywords to retrieve news and \citet{cai_2025} used a hazard event database for a targeted query of news articles and refined results using a Large Language Model (LLM). Our work focuses on the step of \textit{refining} an initial sample of documents retrieved via keyword-matching methods.

\paragraph{Topic Models} TMs such as Latent Dirichlet Allocation (LDA) \citep{lda} and Non-Negative Matrix Factorization (NMF) yield distributions of topics in documents in an unsupervised fashion. Many variations exist, e.g.~keyword-assisted TMs \citep{keyword-tm}, which exploit keywords to guide clustering and circumvent post hoc topic interpretation, CorEx \citep{gallagher-etal-2017-anchored}, which relies on an information-theoretic framework, and Top2Vec \citep{angelov-inkpen-2024-topic}, which performs clustering in an embedding space shared by documents and words. TMs often  aid characterising corpora in climate and ecology research \citep[\textit{inter alia}]{tm-adaptation,stede2023framing,flood-tms,zander2023topic,peura-etal-2025-perspectives,beckles-heidke-2025-thematic,barz-etal-2025-analyzing}. Tuning and assessing the quality of TMs is intricate if performed purely intrinsically \citep{maier2021applying}, but our evaluation is enhanced by annotated data that allows known document properties to be compared to the formed topics. 

\paragraph{TMs for text classification} TMs have been widely used to map texts to classes, e.g.~by feeding their outputs as input features for other classifiers \citep{li2016,anantharaman2019,seifollahi-2021}. Other works aligned topics to classes, either directly \citep{sarioglu-etal-2013-topic}, by experts \citep{hingmire2013} or by configuring the \textit{priors} in a way that induces desired clusters \citep{miller-etal-2016-unsupervised,rubin2012statistical}, e.g.~by relying on relevant keywords \citep{chen2015,zha2019multi,li-xing2016,li2018}. Keyword selection can also derive a lower dimensional set of features for other types of classifier models \citep{onan2016}. \citet{blei-slda} incorporated a response variable into the TM training, to jointly model documents and their classes or scores. While many procedures require adjusting \textit{priors} or the modelling approach, we stick to standard LDA and NMF implementations, which are arguably more accessible for newcomers and researchers from other fields. 

\paragraph{Text classification in climate research} Climate-related text classification is an established NLP task; in many settings, it remains an unsolved problem even for LLMs, with performance often well below 0.75 F1 in the ClimateEval benchmark \citep{kurfali-etal-2025-climateeval}. In the study by \citet{li-etal-2024-using-llms}, a fine-tuned encoder achieved an F1 of 0.98 for identifying relevant documents on climate extreme impacts, but only in English and on a small sample of cleaner Wikipedia entries with climate-related keywords \textit{in their titles}. This restriction likely ensured a majority of relevant matches, but resulted in an unknown number of missed cases. When full texts are considered (as we do), there is less room for false negatives while substantially increasing the need for filtering out false positives, especially with imbalanced datasets.
\\
 
The problem we tackle in this paper is similar to the work by \citet{grasso-etal-2024-nytac}: corpus construction via keyword-based prefiltering and automatic classification. We differ by focusing on German, handling specific hazards separately and exploring TMs for classification, not only for topic analysis as that work did. Our design builds upon existing work with a novel perspective: we do not change the LDA and NMF internal mechanisms and explore the \textit{posterior} probabilities (or normalised scores) they assign to keywords as a means to automatically partition topics and perform binary classification of news about extreme climate events.

%% file: contents/method.tex
This section formalises the task and explains how topic models are applied for binary classification. Then, it describes the two deep learning strategies used for comparison.

\subsection{Task Formalisation}

Let $D$ be a set of documents $d$, each belonging to a binary class $C=\{0, 1\}$, and $V$ be the set of all tokens $w$ that appear in $D$. Class $1$ represents relevant documents. A document classifier is a function $f\colon D \to C$ that maps documents to classes and can be approximated by various methods.  

Furthermore, let $F \subseteq V$ be a set of feature tokens selected from $V$ based on given criteria (e.g.~minimum frequency and part-of-speech tags) and $K \subset F$ be a small set of predefined tokens of interest which we name \textit{keywords}. A trained topic model $M$ with $n$ topics $T$  estimates two distributions: $p_{feat}$, the probability of a feature token in a topic and $p_{topic}$, the probability of a topic in a document. In other words, a document is represented as a probability distribution over topics and a topic, as a probability distribution over feature tokens. 

With $M$'s estimations, we can define a binary relation $R$ between $D$ and $T$ representing whether each document is related to each topic. 
To use $M$ for binary classification, a partition of topics $T$ with two sets is created, each corresponding to a class in $C$. The class of a topic must also correctly classify documents related to such a topic.

\subsection{Classification with Topic Models}

Firstly, a TM is trained on the entire collection of unique documents using selected hyperparameters, following standard procedures (described in Section \ref{sec:experiments}). Then, two further steps are needed: (i) assigning topics to documents and (ii) identifying which topics are to be regarded as relevant. 

For (i), we define the relation $R$ as $p_{topic}(t,d) \geq \theta$ with $0 \geq \theta \geq 1$. That means that if the proportion of a topic $t$ in a document $d$ is at least a threshold, we consider that $d$ discusses topic $t$ (as in Figure \ref{fig:intro-overview}). 

For (ii), we propose two ways to partition topics into two classes, relevant and not relevant, avoiding the usual \textit{post hoc} human interpretation in TMs:

\begin{itemize}
    \item \textbf{keyword proximity}: topic $t$ is assigned to the relevant class if $\exists w\in K: p_{feat} (w, t) > \gamma$. In other words, if the topic assigns a high enough probability to at least one keyword, the topic belongs to the partition of the relevant class.
    
    \item \textbf{top terms}: if there is a keyword among the top $k$ features of a topic (ranked by probability), the topic is assigned to the relevant class.
   
\end{itemize}

The actual classification of each document is made  as follows: if the document is related to at least one of the relevant topics, we consider it to be relevant. Otherwise, it is classified as not relevant.

\subsection{Deep Learning Classifiers}

The performance of our TM approach is compared to two deep learning alternatives: a fine-tuned text embedding model and an LLM. The first is a binary classifier trained using the SetFit framework \citep{setfit} which fine-tunes a pretrained text embedding model with a classification head, aiming at optimising task-specific embeddings based on a set of contrastive examples. The latter prompts an LLM to generate a binary label classifying the document as relevant or not. The implementation details are explained in Section \ref{sec:experiments}.

%% file: contents/data.tex
The data for this study derives from an ongoing project on the collective attention to extreme climate events in the German media. Seven types of hazards were selected (cold waves, droughts, floods, heat waves, landslides, storms and wildfires). The wiso-net news aggregation database\footnote{\url{https://www.wiso-net.de/}} was queried using a pre-defined list of hazard-related keywords, similar to \citep[see Appendix]{li-etal-2024-using-llms,flood-tms,Carvalho2025}. The retrieved collection contained 13,771,411 German news articles from around 370 outlets, spanning from 2000 to 2024, split into separate sub-collections for each extreme climate event. 

\begin{figure}
	\includegraphics[trim={0cm 5.6cm 13.5cm 0},clip,width=\linewidth]{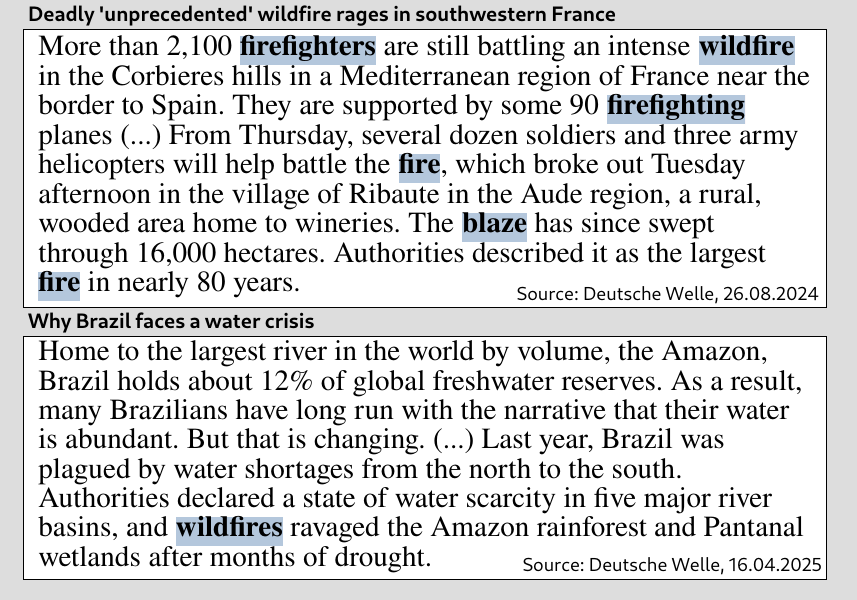}
	\caption{Wildfires as the main topic (top) or mention (bottom) in news excerpts.}
	\label{fig:main-mention}
\end{figure}

\begin{figure*}[ht]
	\centering
	\includegraphics[width=0.44\textwidth]{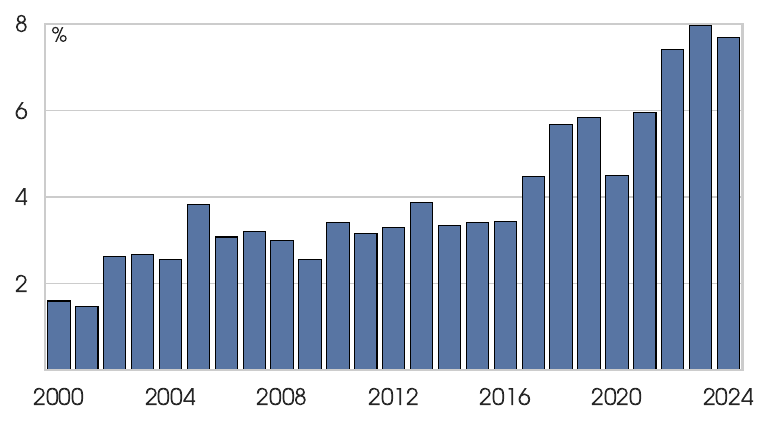}
	\hfill
	\includegraphics[width=0.55\textwidth]{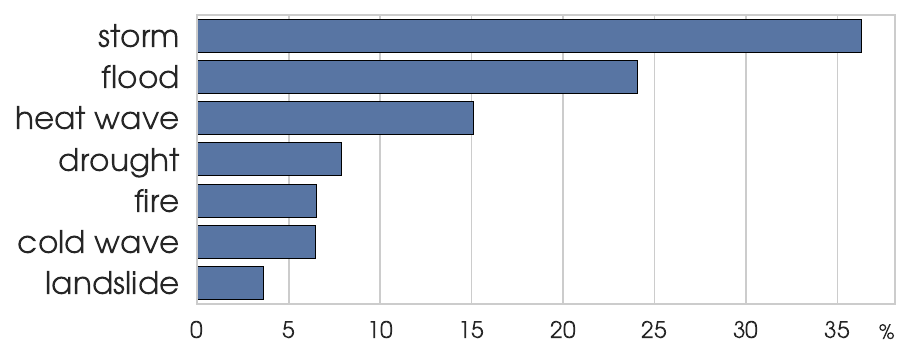}
	\caption{\% of documents per year (left) and \% of documents per extreme climate event (right).}
	\label{fig:data}
\end{figure*}

We make a distinction between two types of relevant news: \texttt{main}, in which the extreme climate event is the main topic, and \texttt{mention}, that refer to it \textit{en passant}, of secondary importance among other more prominent topics, as shown in Figure \ref{fig:main-mention}. Both forms count towards levels of collective attention, but automated identification of the latter is more challenging due to brevity and underspecification.

We are primarily interested in the coverage of \textit{international} events in German news, so basic rule-based procedures were implemented to reduce the amount of local or unrelated news. As the data we are working with was queried via unrestricted keyword matching, many false positives occurred. Of particular relevance to this paper is the filter for what we call ``intruder'' keywords, i.e.~words that derive from valid hazard-related keywords but are unrelated to climate events. For instance, the name \textit{Dürrenmatt}, retrieved by \textit{Dürre} (drought), \textit{Flutlicht} (floodlight), by \textit{Flut} (flood) and \textit{Stürmer} (forward player in football), by \textit{Stürme} (storms). To reduce the number of unrelated documents, we excluded all instances that contained only intruder keywords. Other preprocessing steps for filtering and cleaning the data are summarised in the Appendix.

\paragraph{Final document collection} The previous steps resulted in a sample of 2,438,275 documents (17.71\% of the originally retrieved instances). They have, on average, 537.19 tokens (std$=$362.06) and 28.66 sentences (std$=$20.12). The news database has an inherent temporal bias towards recent years, illustrated in Figure \ref{fig:data} (left). The distribution over types of hazards is also shown in Figure \ref{fig:data} (right). 

\paragraph{Gold standard} A sample of 3,150 documents was randomly selected while ensuring uniform distribution across hazard types (450 each) and years (18 each per hazard) and no duplicates per hazard. Two annotators classified the news as relevant or not (see annotation instructions in the Appendix) while also judging whether the event of interest was the news' main topic or just a mention. Table \ref*{table:annot-stats} shows the percentage of relevant documents identified in the annotated sample. The initial effectiveness of the dictionary-based approach, together with the rule-based preprocessing, depends on the type of extreme climate event: while the landslide portion already reached a precision of almost 0.6, cold and heat waves stayed below 0.2.

Identifying relevant documents is not as straightforward as it may sound. A subset of 100 news was annotated by both annotators independently. The overall agreement proportion in the primary binary decision (relevant or not) was 0.77 ($\kappa = $ 0.53), indicating that there are sources of legitimate disagreement in this decision. Apart from potential errors, disagreements may stem from differences in perception of what constitutes an \textit{extreme} and \textit{concrete} natural event. Some cases which may have involved such subjectivity were (translated from German):

\begin{itemize}
	\item \textbf{cold wave}: ``...the son, travelled on Monday during the \underline{snow chaos} from Cologne to Wismar in order stay by his mother.''
	\item \textbf{drought}: ``the mine was found two weeks ago due to the \underline{low water level} in the Rhine river.''
	\item \textbf{landslide}: ``we knew the situation when a country closed the border or a street was blocked for a week due to a \underline{landslide} or something else.''
\end{itemize}

\begin{table}[t]
	\centering
	
	\begin{tabular}{rrr}
		\toprule
		& relevant & main topic \\
		\midrule
		cold wave & 14.44 & 4.22 \\
		drought & 36.00 & 4.22 \\
		flood & 43.33 & 12.44 \\
		heat wave & 19.78 & 2.44 \\
		landslide & 58.67 & 17.56 \\
		storm & 27.56 & 7.33 \\
		wildfire & 41.56 & 21.11 \\
		\cline{1-3}
		\bottomrule
	\end{tabular}
	
	\caption{Percentage of relevant documents for each type of extreme climate event in the gold standard.}
	\label{table:annot-stats}
\end{table}

%% file: contents/experiments.tex
The varying estimated proportions of relevant documents for each hazard sample suggest that these phenomena manifest differently not only in their nature but also in their coverage and linguistic features. Therefore, each classification strategy was conducted for each type of extreme climate event separately. The annotated sample was randomly split into a training and a test set with 350 and 100 instances, respectively, for each hazard. The presented results were computed in the test split.\footnote{Note that the use of train/test splits depends on the classifier. TMs's unsupervised fitting included all unique documents, since the objective here is not to generalise to unseen data but to optimise for topics that best fit our own documents. Still, only the train split was used to select the best model configuration to avoid overfitting to the test data in this choice. The text embeddings model used the train split for fine-tuning. The LLM was directly prompted with the test data in a zero-shot approach.}

\paragraph{Topic models} Documents were preprocessed to extract their features partially based on the procedure by \citet{grasso-etal-2024-nytac}. We used Spacy's\footnote{\url{https://spacy.io/}} model \texttt{de\_core\_news\_lg} to tokenize, lemmatise and label tokens with their part-of-speech tags. Tokens with less than 3 characters and stopwords were removed, as well as non-alphabetical characters. All tokens were lowercased. The feature selection involved two criteria: the term's document frequency and part-of-speech tag. All keywords were kept as features, even if they did not meet the minimum frequency threshold, to ensure they had a chance to contribute to forming a topic. To avoid the induction of topics based on duplicated news, only one instance of texts with high similarity was included. Gensim's\footnote{\url{https://radimrehurek.com/gensim/}} implementation of the LDA and NMF methods was used to train topic models. The number of topics was a hyperparameter. For LDA, the \texttt{eta} and \texttt{alpha} arguments were set to \texttt{auto}. We run various combinations of the three hyperparameters (minimum document frequency, part-of-speech tags and number of topics) and, for each model, we computed results varying the values for $k$, $\gamma$, for top term and keyword proximity, and $\theta$. For each hazard, we selected the best-performing models in the training split. Specific parameters and the final configuration that produced the results are in the Appendix. The code is available at \url{https://codeberg.org/briemadu/tm-as-classifier}.

\paragraph{Text embeddings} This classifier was trained via the Small-Text \citep{schroder-etal-2023-small} wrapper implementation around SetFit (with its default configuration in HuggingFace) and Sentence Transformers \citep{reimers-gurevych-2019-sentence}. In this method, the classification is performed by a logistic regression component on top of the fine-tuned text embeddings. We opted for the \texttt{BAAI/bge-m3} text embeddings released by \citep{chen-etal-2024-m3} due to the model's multilingual capabilities and longer context length (8,192 tokens), since standard Sentence Transformers that typically allow only up to 512 tokens would not suffice for longer news articles. Training was performed with a batch size of 16 instances and a learning rate of $10^{-5}$. 

\paragraph{LLM} Since the purpose of this paper is not to benchmark LLM performance, we chose only one model to serve as a reference. Results were produced by \texttt{mistralai/ministral-3-14b-reasoning}.\footnote{ \url{https://huggingface.co/mistralai/Ministral-3-14B-Reasoning-2512}}  We selected an open-weight model that could be run locally and keep the data in our own infrastructure.\footnote{We did not compare results to closed commercial models as they are at odds with open science principles.} The prompt contained instructions similar to those given to the annotators, including the definition of the hazard and of the labels, the hazard's keywords and the main body of the news article. The exact prompt and values are in the Appendix. We had to programatically parse answers that included spurious prefixes before the actual label.

\subsection*{Evaluation}

The models' performance was quantitatively assessed with conventional binary classification metrics: precision, recall and F1 score of the positive class. The test sample's precision and a presumed recall of 1 were used as a baseline to measure how much the classifiers improve retrieval precision without reducing its recall. The evaluation was enriched with a detailed analysis of the TM results.

We present results for three variations of TMs: \textsc{tm-f1} was run with the configuration that resulted in the highest F1 score (on the training split) in our hyperparameter search; \textsc{tm-b} uses the configuration that balanced precision and recall to be both as high as possible; and \textsc{tm-p} has the configuration with the highest precision while retaining some level of recall. We also compare results to an ensemble strategy that performs classification via majority voting across the outputs of \textsc{tm-b}, fine-tuned text embeddings and LLM classifiers.

%% file: contents/results.tex
\begin{table}[t]
	\centering
	\begin{tabular}{rccc |c}
		\toprule
		& P & R & F1 &  $n$ \texttt{main}\\
		\midrule
		baseline & 0.350 & 1.000 & 0.519 & 58 \\
		\cmidrule{1-5}
		\textsc{tm-f1} & 0.637 & 0.710 & 0.672 & 56 \\
		\textsc{tm-b} & \textbf{0.710} & 0.649 & 0.678 & 55 \\
		\textsc{tm-p} & \textbf{0.808} & 0.396 & 0.532 & 47 \\
		fine-tuned & 0.647 & 0.853 & 0.736 & 57 \\
		llm & 0.583 & \textbf{0.976} & 0.730 & 58 \\
		majority & 0.701 & 0.890 & \textbf{0.784} & 58 \\
		\cline{1-5}
		\bottomrule
	\end{tabular}
	
	\caption{Aggregated results: binary precision, recall and F1 score of all classifiers in the test split and the number of news of type \texttt{main} correctly identified.}
	\label{table:overall}
\end{table}

\begin{table}[h!]
	\centering 
	\small
	\begin{tabular}{llccc}
		\toprule
		&  & P & R & F1 \\
		\midrule
		\multirow[t]{5}{*}{cold wave} & baseline & 0.170 & 1.000 & 0.291 \\
		\cmidrule{2-5}
		& \textsc{tm-f1} & 0.471 & 0.471 & 0.471 \\
		& \textsc{tm-b} & \textbf{0.583} & 0.412 & 0.483 \\
		& \textsc{tm-p} & 0.500 & 0.059 & 0.105 \\
		& fine-tuned & 0.297 & 0.647 & 0.407 \\
		& llm & 0.455 & \textbf{0.882} & 0.600 \\
		& majority & 0.542 & 0.765 & \textbf{0.634} \\
		
		\cmidrule{1-5}
		\multirow[t]{5}{*}{drought} & baseline & 0.440 & 1.000 & 0.611 \\
		\cmidrule{2-5}
		& \textsc{tm-f1} & 0.517 & 0.682 & 0.588 \\
		& \textsc{tm-b} & 0.622 & 0.523 & 0.568 \\
		& \textsc{tm-p} & \textbf{0.938} & 0.341 & 0.500 \\
		& fine-tuned & 0.686 & 0.795 & 0.737 \\
		& llm & 0.525 & \textbf{0.955} & 0.677 \\
		& majority & 0.692 & 0.818 & \textbf{0.750} \\
		
		\cmidrule{1-5}
		\multirow[t]{5}{*}{flood} & baseline & 0.360 & 1.000 & 0.529 \\
		\cmidrule{2-5}
		& \textsc{tm-f1} & 0.605 & 0.639 & 0.622 \\
		& \textsc{tm-b} & 0.595 & 0.611 & 0.603 \\
		& \textsc{tm-p} & \textbf{0.750} & 0.167 & 0.273 \\
		& fine-tuned & 0.737 & 0.778 & 0.757 \\
		& llm & 0.600 & \textbf{1.000} & 0.750 \\
		& majority & 0.738 & 0.861 & \textbf{0.795} \\
		
		\cmidrule{1-5}
		\multirow[t]{5}{*}{heat wave} & baseline & 0.200 & 1.000 & 0.333 \\
		\cmidrule{2-5}
		& \textsc{tm-f1} & 0.423 & 0.550 & 0.478 \\
		& \textsc{tm-b} & \textbf{0.600} & 0.450 & 0.514 \\
		& \textsc{tm-p} & \textbf{0.600} & 0.150 & 0.240 \\
		& fine-tuned & 0.439 & 0.900 & \textbf{0.590} \\
		& llm & 0.322 & \textbf{0.950} & 0.481 \\
		& majority & 0.429 & 0.900 & 0.581 \\
		
		\cmidrule{1-5}
		\multirow[t]{5}{*}{landslide} & baseline & 0.580 & 1.000 & 0.734 \\
		\cmidrule{2-5}
		& \textsc{tm-f1} & 0.785 & 0.879 & 0.829 \\
		& \textsc{tm-b} & 0.778 & 0.845 & 0.810 \\
		& \textsc{tm-p} & 0.816 & 0.690 & 0.748 \\
		& fine-tuned & \textbf{0.877} & \textbf{0.983} & \textbf{0.927} \\
		& llm & 0.826 & \textbf{0.983} & 0.898 \\
		& majority & 0.826 & \textbf{0.983} & 0.898 \\
		
		\cmidrule{1-5}
		\multirow[t]{5}{*}{storm} & baseline & 0.270 & 1.000 & 0.425 \\
		\cmidrule{2-5}
		& \textsc{tm-f1} & 0.680 & 0.630 & 0.654 \\
		& \textsc{tm-b} & \textbf{0.800} & 0.593 & 0.681 \\
		& \textsc{tm-p} & 0.625 & 0.185 & 0.286 \\
		& fine-tuned & 0.558 & 0.889 & 0.686 \\
		& llm & 0.614 & \textbf{1.000} & 0.761 \\
		& majority & 0.714 & 0.926 & \textbf{0.806} \\
		
		\cmidrule{1-5}
		\multirow[t]{5}{*}{wildfire} & baseline & 0.430 & 1.000 & 0.601 \\
		\cmidrule{2-5}
		& \textsc{tm-f1} & 0.773 & 0.791 & 0.782 \\
		& \textsc{tm-b} & 0.825 & 0.767 & 0.795 \\
		& \textsc{tm-p} & \textbf{0.844} & 0.628 & 0.720 \\
		& fine-tuned & 0.750 & 0.837 & 0.791 \\
		& llm & 0.662 & \textbf{1.000} & 0.796 \\
		& majority & 0.809 & 0.884 & \textbf{0.844} \\
		
		\cline{1-5}
		\bottomrule
	\end{tabular}
	
	\caption{Detailed results: binary precision, recall and F1 score of all classifiers in the test split for each hazard type.}
	\label{table:detailed-results}
\end{table}

\paragraph{Aggregated results} We first examine results aggregated over the whole test split ($n=700$), i.e.~including all extreme climate events. Table \ref{table:overall} shows precision, recall and F1 score for all classifiers. The rightmost column shows the number of news articles of type \texttt{main} that were correctly identified as relevant. All classifiers succeeded in considerably increasing the low proportion of relevant documents in the keyword-based sample, but TMs and deep learning strategies behaved differently in how precision and recall were balanced. While the LLM had almost maximum recall with a substantial margin over other models, its precision was the lowest. \textsc{tm-p} had the highest precision but at the cost of low recall. \textsc{tm-b} achieved the second highest precision with a more reasonable recall. The majority voting approach led to the highest F1 score. If we focus on the identification of news of type \texttt{main}, all classifiers (apart from \textsc{tm-p}) performed very well, identifying at least 55 out of the 58 instances.

\paragraph{Results by hazard} Aggregated results can mask variations in performance for each underlying hazard. Table \ref{table:detailed-results} summarises results by hazard type, in line with the fact that models were trained separately. We can see that metrics varied greatly depending on the phenomenon: the lowest best F1 score of 0.59 occurred for heat wave whereas the highest best of 0.92 was observed for landslide. Majority voting achieved the best F1 scores for five hazards and the fine-tuned text embeddings for the other two. The LLM consistently held the highest recall in all hazards. \textsc{tm-b} had the best precision in the three most imbalanced (cold waves, heat waves and storms).

\begin{figure*}[h!]
	\includegraphics[trim={0cm 0cm 0cm 0},clip,width=\linewidth]{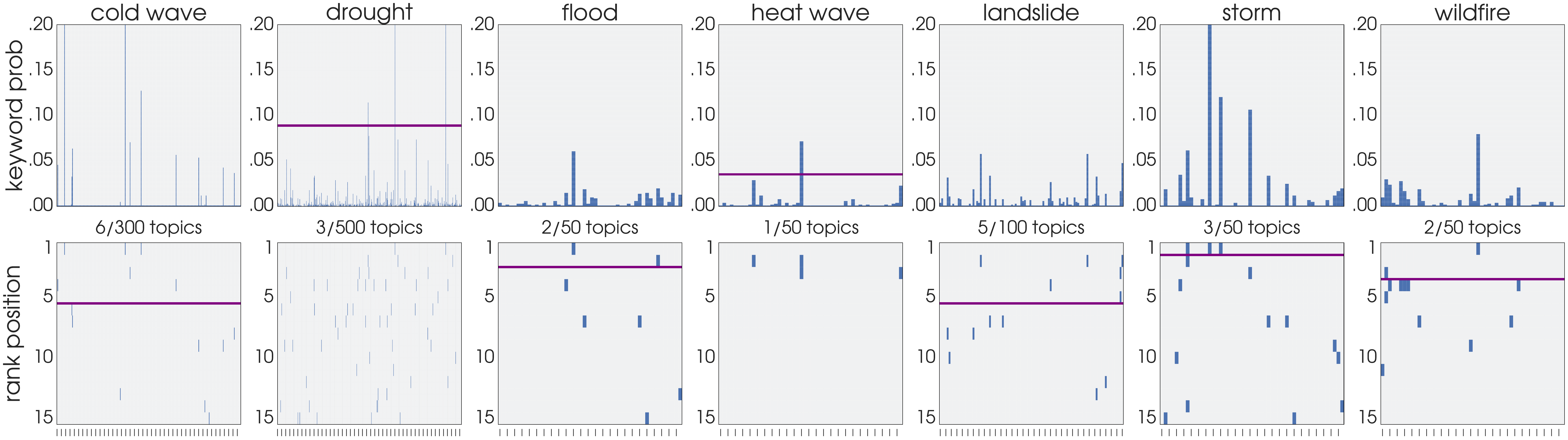}
	\caption{High-level overview of how relevant topics were identified by \textsc{tm-b}. In the first row, bars represent the maximum probability each topic (x-axis) assigns to keywords. The second row highlights in dark the positions in the rank that contain keywords in each topic (x-axis). The purple horizontal bars represent the optimal $\gamma$ and $k$, respectively, in our experiments (see Appendix for exact numerical values).}
	\label{fig:thresholds}
\end{figure*}

\paragraph{Discussion} 
In aggregated results, TM performance was indeed lower than that of deep learning strategies, but the moderate reduction of only around 0.06 in F1 score still provides a much-desired gain in interpretability: we can explain exactly why each document was classified as relevant. The deep learning strategies tended to incur more false positives whereas TMs could reduce the proportion of unrelated documents while causing more false negatives. Models with higher precision but low recall, like \textsc{tm-p}, can still be useful when sample precision is a priority, since a sample with low recall may still be representative and of enough size in large datasets. High precision helps reduce the impact of unrelated documents in downstream analyses. For situations in which news of type \texttt{main} are more important than \texttt{mention}, the two best TMs and deep learning classifiers worked in like manner, not missing the vast majority of instances. 

The majority voting approach seemed to draw out advantages from each model, achieving the best F1 score. Still, employing three computationally costly models for this task is hardly justifiable in practice, given the modest overall increase in the aggregated F1 score compared to single models. 

Classification of news turned out to be hazard-dependant. There was no one-size-fits-all best solution across all hazards. The fine-tuned text embeddings balanced precision and recall well in general, but in three hazards (cold waves, storms and wildfires) the F1 score of the TM approaches was on par or better than it, which is a very interesting finding given that TMs are unsupervised and do not rely on the currently prevailing deep learning paradigm. Landslides and wildfires were the easiest to identify with all metrics above 0.75 (except for the LLM's precision for wildfire) in all models apart from \textsc{tm-p}. Cold and heat waves were the most challenging with suboptimal results even for the majority voting method. 

Note, however, that comparisons between models should be done with caution, as these experimental estimates by hazard type were computed from samples of only 100 documents each. Rare events become very sensitive to individual predictions in such a small sample. For instance, cold waves contain only 17 relevant documents on which to measure precision and recall, so that a single swapped prediction by a model would already cause a 5.8\% increase or decrease in recall.

%% file: contents/analysis.tex
\begin{figure}[h!]
	\includegraphics[trim={0cm 6cm 8.5cm 0},clip,width=\linewidth]{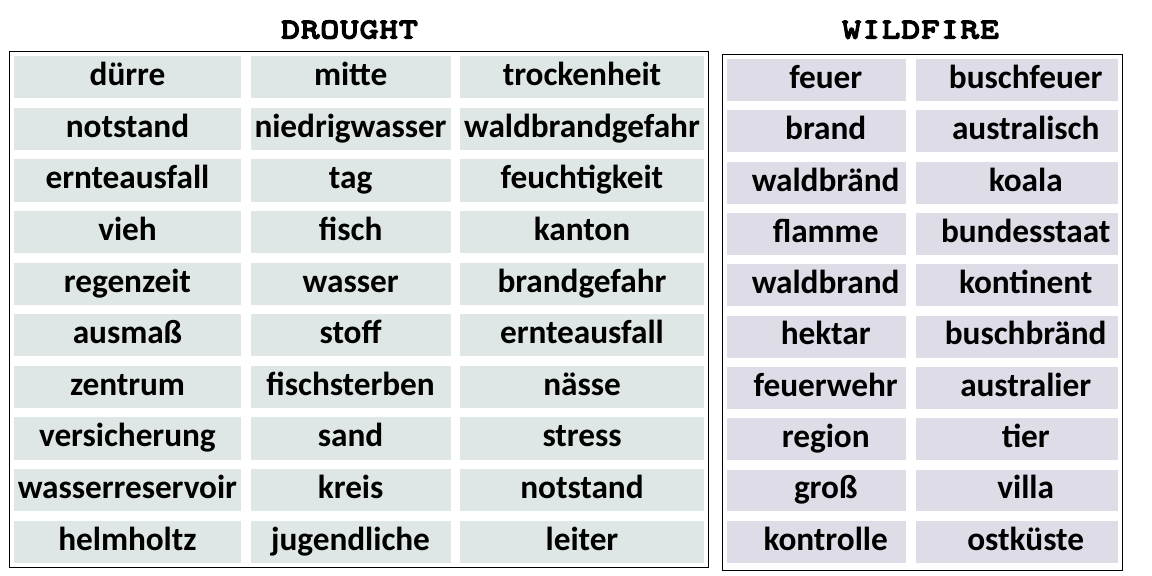}
	\caption{Top 10 terms in each of the topics considered as relevant for drought and wildfire.}
	\label{fig:topterms}
\end{figure}

In this section, we explore TM's interpretability by providing more details on the \textsc{tm-b} models' behaviour. In our non-exhaustive hyperparameter search, LDA achieved the best performance for six hazard types, while NMF was superior only for drought. The optimal thresholds $\theta$ for assigning topics to documents were between 0.028 and 0.076. Figure \ref{fig:thresholds} illustrates how relevant topics were selected. The top terms decision method achieved the highest performance across five hazards with $k$ values ranging from 1 to 5. Keyword proximity was superior only for drought and heat wave using $\gamma=$0.09 and 0.036, respectively. The number of selected topics for each hazard varied from 1 to 6.

\begin{figure}[h!]
	\includegraphics[trim={0cm 0cm 0cm 0},clip,width=\linewidth]{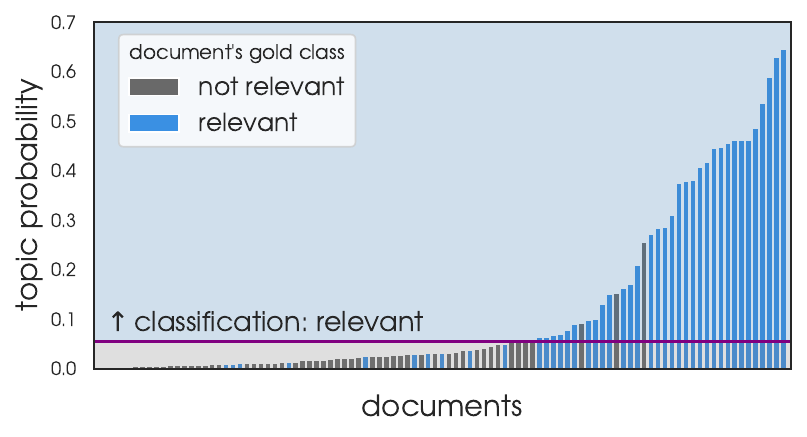}
	\hfill
	\caption{Example of the $\theta$ threshold for assigning a relevant fire topic to documents. Above the purple horizontal line, documents are classified as relevant, with a few mistakes with respect to the gold standard.}
	\label{fig:theta}
\end{figure}

Here we focus on wildfires and droughts as they had the smallest and largest differences in F1 score, respectively, in relation to deep learning strategies. For wildfires, topics were considered relevant if a hazard-related keyword was among their top 3 most probable terms. That resulted in 2 out of 50 topics being considered as relevant. For drought, keyword proximity selected 3 out of 500 topics as relevant. The top 10 lemmas representing these topics are shown in Figure \ref{fig:topterms}. Figure \ref{fig:theta} illustrates the effect of $\theta$ for wildfire's leftmost relevant topic:
how well (not) relevant documents are classified based on the $\theta$ parameter for the rightmost relevant topic in fire: documents with topic probability above $\theta$ are classified as relevant, with a few wrong predictions.  

\paragraph{Wildfires} Out of the 100 test instances, 83 were correctly classified. We inspected the 17 errors. The 7 false positives are texts that do refer to fires, but they are either not wildfires (e.g.~fires in houses or industrial plants) or discuss technologies to combat wildfires. One of the documents describes a wildfire warning that also refers to an extinguished fire, which may have been missed by the annotators. Among the 10 false negatives, most contain mentions to wildfires occurring in discussion about other main topics (financial markets, heat waves, conferences) in documents that are a concatenation of various news articles (a problem we inherited from the original database were not able to fully solve automatically). In such cases, the fire-related topic may not have been salient enough to be assigned to the document. Although two topics were selected as relevant, only the left one was responsible for making all predictions on the test set. The second topic appeared in only one text, indicating that, although potentially relevant, it may have specialised too much during training. All documents of type \texttt{main} were correctly identified.

\paragraph{Drought} For this hazard, 65 documents were correctly classified, with 14 false positives and 21 false negatives. Many false positives treat drought as a broad phenomenon rather than a concrete event, for example, when discussing drought-tolerant plants, vegetation stress, or climate change.
Such cases are difficult to distinguish using TMs and may also reflect ambiguities in the annotation. False negatives show no clear patterns. Since this model relied on 500 topics, the drought concept may have been diffuse across multiple topics: in Figure \ref{fig:thresholds}, it is evident that drought keywords appear in various topics not selected as relevant. The only \texttt{main} document that was incorrectly classified has only one keyword, a compound noun (\textit{Dürregebieten}), which was not included as a feature. Including all words \textit{containing} keywords as features could have prevented this but it introduces additional noise from low-frequency terms that form topics.

%% file: contents/discussion.tex
This work was primarily motivated by the lack of a comprehensive global database of extreme climate disasters. Existing disaster databases, for instance the EM-DAT \citep{Delforge2025}, are shaped by reporting practices and inclusion thresholds (e.g.~at least 10 fatalities), which have been widely discussed for their biased coverage toward large, well-documented events and wealthier regions, systematically under-representing some regions and hazards \citep{Jones2023}. Our method is designed to support bottom-up data-driven analyses by bypassing the inherent incompleteness and structural biases of top-down lists of worldwide extreme climate events \citep{gall2009}. Our procedure permits the inclusion of news about events that did not meet the arbitrary inclusion criteria of disaster databases. 

Rather than claiming the superiority of one model type for news classification, we have provided evidence that the results are hazard-dependent. This is an important finding for climate impact and adaptation research: the way different hazards are reported in the news varies, so solutions that treat all climate-related hazards as a single category (i.e.~disasters in general) risk masking important performance variations, leading subsequent conclusions to be biased towards those that are easier to identify. In this context, an advantage of our approach is that we explicitly consider hazards separately, enabling more reliable downstream analyses.

The exact reasons for such differences require further investigation. First, each hazard is inherently distinct in the abruptness of its onset, its duration, its frequency, and its perceived severity. Then, media coverage can differ depending on socio-economic and geo-political factors. Finally, there is linguistic and discourse-related variation. For instance, while some keywords are very specific to climate events, others are polysemic and appear in multi-word expressions. The interdependencies among these layers are worth studying. Some events are hard to pinpoint even for humans, which can impact gold standards. Treating extreme climate events as a monolithic concept is thus not advisable in NLP tasks. Besides, since multi-hazard events occur in reality, another promising way forward is to analyse how they also co-occur in news.

We aimed to reduce human input in TM interpretation by selecting thresholds automatically and minimizing hyperparameter choices. Further work can investigate whether manual selection of keywords and topics can improve results. Our preliminary experiments with CorEx and Top2Vec yielded comparable results, so we prioritised the more traditional LDA/NMF methods in this study. However, other TM variations can be further investigated, including tuning priors to promote clearer keyword-related topics. The fine-tuned text embeddings achieved some of the highest F1 scores using only 350 documents and can potentially be further improved with active learning  \citep{schroder2020surveyactivelearningtext}.

LLMs are being uncritically employed for many NLP tasks. We have shown that even a model with 14b parameters was not sufficiently precise. Our results add to the evidence that LLMs require careful evaluation as any other model. If LLM-based approaches are to be used, TMs can still be helpful in shrinking the amount of unrelated documents (e.g.~by excluding those that have high probability for totally unrelated topics), thereby reducing the considerable environmental and financial costs of using LLMs.

%% file: contents/conclusions.tex
We have presented a comparative analysis of three binary classifiers for refining collections of news articles on extreme climate events retrieved via keyword-based approaches. Although the LLM and the fine-tuned text embeddings had a higher F1 score in general, the drop in comparison to TMs was 0.148 on the worst case (drought) but also only 0.001 on the best case (wildfire). This is remarkable given TMs' unsupervised training and the simplicity of the keyword-guided topic selection process. Depending on the use case, this difference may be acceptable given other advantages, such as higher precision. Besides, the reason for deep learning-based predictions are beyond human comprehension, whereas decisions based on TMs are fully transparent and explainable.

%% file: contents/limitations.tex
The rule-based filtering may have excluded relevant documents, although it was a price worth paying to reduce the immense volume of unrelated news and to keep the task computationally tractable. Although we are seeking to identify extreme climate events, other types of disasters (.e.g.~urban fires and industrial accidents that cause dam collapse) could not yet be fully distinguished by our methods.  

The test samples for each extreme climate event contain only 100 documents each, which may obscure variance in the estimates. More definitive claims about differences in models' behaviour require cross-validation and, ideally, a larger sample. The performance of the classifiers is bounded by the quality of the annotation. Despite best efforts, ambiguity is not always easy to resolve and arbitrary decisions can impact models' training and evaluation. 

We presented results for varying TM set-ups as we opted for selecting the best-performing configurations. Still, keeping it constant would facilitate the direct comparison across hazards. The hyperparameter search for TM considered only a few dozen combinations of the number of topics, POS-tags and minimum document frequency. This can potentially be further refined for each hazard separately. 

We did not perform extensive prompt engineering for the LLM, as these models are supposed to parse natural language instructions well; still, given their unpredictable nature, minor changes to the prompt might have led to different outcomes. Larger models may yield better results, but our focus here was on lower-scale, local solutions.

%% file: contents/appendix.tex
\subsection*{Further details about preprocessing and filtering}

The lists of German keywords used for each hazard are shown in Figures \ref{fig:prompts1}, \ref{fig:prompts2}, and \ref{fig:prompts3}. Although \textit{derecho} was included as a keyword initially, texts containing only this keyword were removed in a preprocessing step. The inclusion of \textit{Regenfälle} (rainfalls) as a keyword for flood resulted in the inclusion of some texts that may not be about floods. 

Here, we provide a summary of the preprocessing and filtering steps applied to the original document collection. Exact implementation details are documented in the preprocessing code which is available upon request.

We removed exactly duplicated instances, i.e.~those pairs or groups of documents for which \textit{all} metadata values were exactly the same. Documents with the same text but published by different outlets or on different dates were kept as they count separately towards media attention.

The regex pattern \texttt{'<.*?>'} was used to remove reminiscent html content. To split (at least part of) the documents that have been concatenated as a single instance, despite being composed of several different pieces of news, we used another regex pattern with frequent news agencies abbreviations (e.g.~dpa and afp) that often appeared in parentheses in between such concatenations. 

Approximately duplicated texts were identified using the MinHash algorithm to estimate Jaccard similarity, with a threshold of 0.8 slightly more conservative than the empirical choice in \citep{MadrugadeBrito2024}. This was not used to exclude any document, but helped ensure that annotators did not annotate the same text (from different news sources) for the same hazard and that TMs were not trained on similar texts that would form spurious clusters. 

Spacy's German model \texttt{de\_core\_news\_lg} was used to parse each text and retrieve tokens and sentence counts. 

We also applied filters to reduce  the number of unrelated documents and local news. The inclusion criteria were as follows:

\begin{itemize}
	\item The document contains at least one keyword related to its assigned hazard. Although this was an imposed criterion for the database query, after splitting concatenated documents, were a few cases of texts that no longer contained keywords.
	\item The document contains at least one keyword of its assigned hazard which is \textit{not} an intruder.
	\item The document's outlet reportedly belongs to the German press.
	\item In case of exact duplicates (regarding all fields), only one instance was kept.
	\item The document's ressort does not contain the word \texttt{lokal}, since we are only interested in international extreme climate events.
	\item The number of tokens is at least 30 and no more than 1,700. The thresholds were selected based on empirical observations of the distribution's histogram and by taking into account an initial batch of annotated documents.
	\item The document contains at least one of the following: a country name, a nationality (as an adjective or a noun) or a city name.
	\item The first token of the document is not the name of a German city followed by a full stop.
	\item The proportion of non-alphabetical characters is less than 0.11. The threshold was selected based on empirical observations of the distributions, also considering an initial batch of annotated documents.
\end{itemize}

\subsection*{Further details about the annotation}

\begin{figure*}
	\centering
	\fbox{\includegraphics[width=0.85\linewidth]{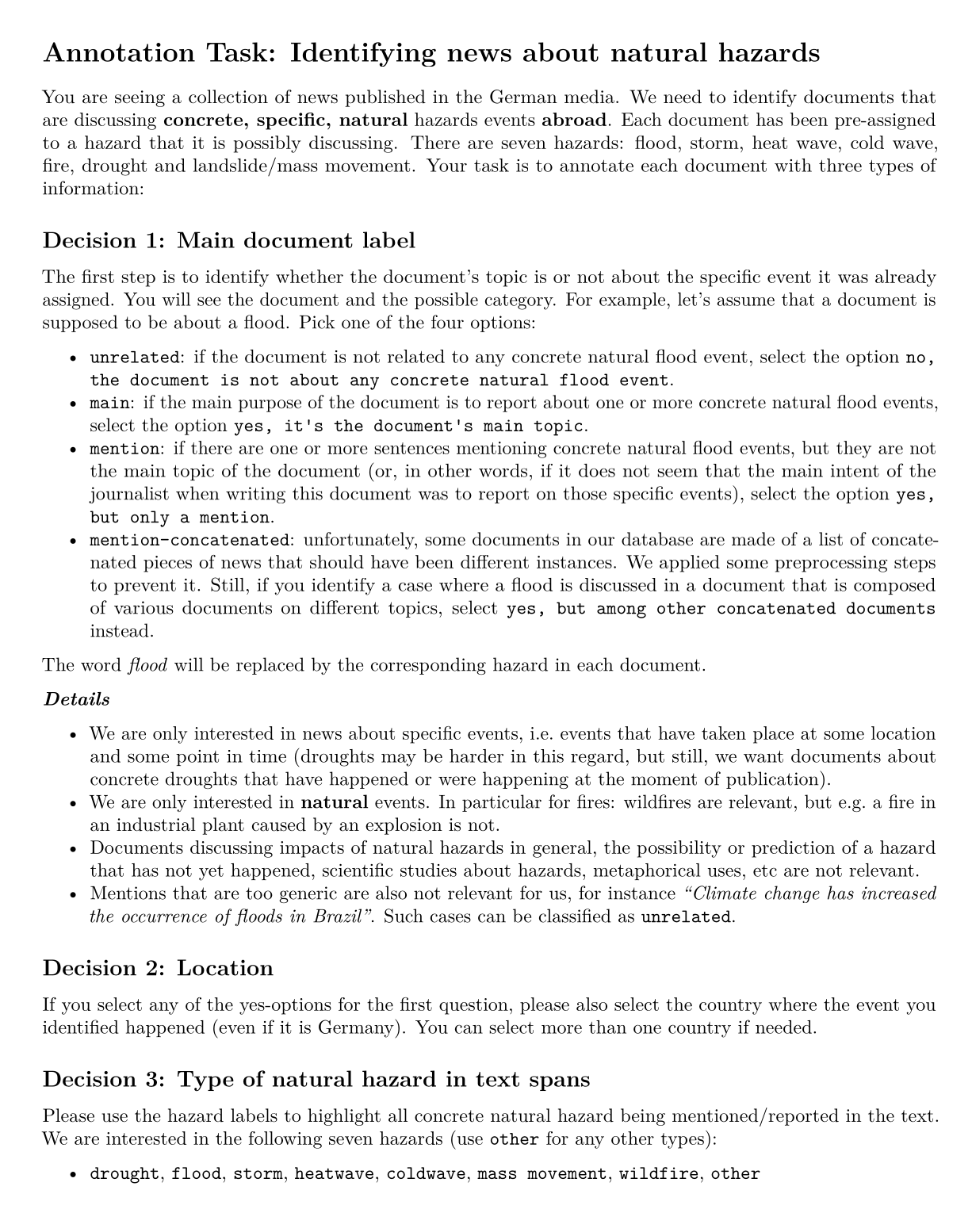}}
	\caption{Instructions given to the annotators.}
	\label{fig:annot-instructions}
\end{figure*}

Figure \ref{fig:annot-instructions} shows the instructions given to the two annotators. They also identified the sentences that refer to each type of hazard and the country where it occurred. These variables will be used in future studies.

\subsection*{Further details about the classifiers}

\paragraph{LLM} Figures \ref{fig:prompts1}, \ref{fig:prompts2}, and \ref{fig:prompts3} show the values used to fill in the hazard-dependent slots in the prompt for the LLM, which is shown in Figure \ref{fig:llm-prompt}. Definitions were translated from the EM-DAT' glossary\footnote{\url{https://doc.emdat.be/docs/data-structure-and-content/glossary/}}, except for storm, not defined by EM-DAT, for which we used Wikipedia\footnote{\url{https://en.wikipedia.org/wiki/Storm}}.

\paragraph{Topic Models} The number of iterations and passes were fixed at 400 and 20, respectively. The random seed was set to 123. Table \ref{tab:tm-docs} shows the number of documents used to train the TMs for each extreme climate event, i.e.~the unique texts in the collection. Table \ref{tab:tm-config} shows the selected hyperparameters for the topic model configuration of each extreme climate event. In Table \ref{tab:tm-terms} we show all topics selected as relevant for \textsc{tm-b} characterised by their top 10 terms with the highest probability.

\vspace{1cm}
\begin{table}[h]
	\centering
	\begin{tabular}{rr}
		\toprule
		cold & 91,140 \\
		drought & 96,441 \\
		flood & 334,607 \\
		heat & 197,800 \\
		landslide & 45,039 \\
		storm & 488,068 \\
		fire & 78,426 \\
		\cline{1-2}
		\bottomrule
	\end{tabular}
	\caption{Number of documents used to train the topic models for each hazard.}
\label{tab:tm-docs}
\end{table}

\begin{table*}[t]
	\centering
	\footnotesize
	
	\begin{tabular}{llllrrrlr}
		\toprule
		&  & model & decision method & $\theta$ & $\gamma$ or $k$ & min.~doc freq & tags & topics \\
		\midrule
		\multirow[t]{7}{*}{\textsc{tm-f1}} & landslide & lda & top terms & 0.058 & 5.000 & 100 & noun, verb, adj & 100 \\
		& fire & lda & top terms & 0.040 & 3.000 & 50 & noun, verb, adj & 50 \\
		& flood & lda & top terms & 0.060 & 2.000 & 1000 & noun, verb, propn & 50 \\
		& storm & lda & top terms & 0.024 & 2.000 & 10000 & noun, verb, adj & 50 \\
		& drought & nmf & keyword proximity & 0.016 & 0.198 & 500 & noun & 500 \\
		& heat & lda & keyword proximity & 0.024 & 0.108 & 100 & noun, verb, propn & 100 \\
		& cold & lda & top terms & 0.024 & 5.000 & 500 & noun, verb, adj & 300 \\
		\cmidrule{1-9}
		\multirow[t]{7}{*}{\textsc{tm-b}} & landslide & lda & top terms & 0.076 & 5.000 & 100 & noun, verb, adj & 100 \\
		& fire & lda & top terms & 0.054 & 3.000 & 50 & noun, verb, adj & 50 \\
		& flood & lda & top terms & 0.062 & 2.000 & 1000 & noun, verb, propn & 50 \\
		& storm & lda & top terms & 0.030 & 1.000 & 10000 & noun, verb, adj & 50 \\
		& drought & nmf & keyword proximity & 0.028 & 0.090 & 500 & noun & 500 \\
		& heat & lda & keyword proximity & 0.052 & 0.036 & 100 & noun, verb, adj & 50 \\
		& cold & lda & top terms & 0.028 & 5.000 & 500 & noun, verb, adj & 300 \\
		\cmidrule{1-9}
		\multirow[t]{7}{*}{\textsc{tm-p}} & landslide & lda & keyword proximity & 0.064 & 0.054 & 50 & noun, verb, adj & 100 \\
		& fire & lda & top terms & 0.064 & 1.000 & 500 & noun, propn & 100 \\
		& flood & lda & top terms & 0.054 & 1.000 & 1000 & noun, verb, propn & 50 \\
		& storm & lda & top terms & 0.120 & 5.000 & 10000 & noun, verb, adj & 50 \\
		& drought & nmf & top terms & 0.034 & 1.000 & 500 & noun, verb, adj & 500 \\
		& heat & lda & top terms & 0.148 & 2.000 & 500 & noun, verb, propn & 50 \\
		& cold & lda & keyword proximity & 0.146 & 0.036 & 5000 & noun, verb, adj & 50 \\
		\cline{1-9}
		\bottomrule
	\end{tabular}
	
	\caption{Hyperparameters of all topic models that produced the presented results.}
	\label{tab:tm-config}
\end{table*}

\begin{figure*}
\begin{lstlisting}[language=json,lastline=20]
landslide: 

    hazard: Erdrutsch
            
    keywords: Erdrutsch, Felssturz, Felsstürz, Schlammlawine, Massenbewegung, Hangrutsch, Hangbewegung, Rutschung, Bodenrutsch, Hangabrutschung, Murgang, Gerölllawine, Rutschhang, Rutschhäng, Rutschgefahr, Felslawine, Mure
    
    hazard_event: Erdrutschereignisse
    
    definition: Jede Art von mäßiger bis schneller Bodenbewegung, einschließlich Lahare, Schlammlawinen und Murgänge (unter trockenen/nassen Bedingungen). Ein Erdrutsch ist eine durch die Schwerkraft gesteuerte Bewegung von Erde oder Gestein, deren Geschwindigkeit in der Regel zwischen langsam und schnell liegt, jedoch nicht sehr langsam ist. Er kann oberflächlich oder tief sein, aber das Material muss eine Masse bilden, die einen Teil des Hangs oder den Hang selbst ausmacht. Die Bewegung muss nach unten und nach außen mit einer freien Fläche erfolgen. ODER Jede Art von Abwärtsbewegung von Erdmaterialien unter hydrologisch trockenen Bedingungen. ODER Arten von Massenbewegungen, die auftreten, wenn starker Regen oder schnelle Schnee-/Eisschmelze große Mengen an Vegetation, Schlamm oder Gestein unter dem Einfluss der Schwerkraft einen Hang hinunterbefördern.
    
fire:
    
    hazard: Waldbrand
    
    keywords: Flächenbrand, Flächenbränd, Waldbrand, Waldbränd, Wildfeuer, Landschaftsbrand, Landschaftsbränd, Buschfeuer, Vegetationsbrand, Vegetationsbränd, Naturbrand, Naturbränd, Großbrand, Großbränd, Forstbrand, Forstbränd, Heidebrand, Heidebränd
    
    hazard_event: Waldbrandeereignisse
    
    definition: Jede unkontrollierte und nicht vorgeschriebene Verbrennung oder das Abbrennen von Pflanzen in einer natürlichen Umgebung wie Wald, Grasland, Buschland oder Tundra, die natürliche Brennstoffe verbraucht und sich aufgrund von Umweltbedingungen (z. B. Wind oder Topografie) ausbreitet. Waldbrände können durch Blitzeinschläge oder menschliches Handeln ausgelöst werden.
    
\end{lstlisting}
	\caption{Keywords and values used in the prompts for each hazard (1/3).}
	\label{fig:prompts1}
\end{figure*}

\begin{figure*}
\begin{lstlisting}[language=json,firstline=21,lastline=50]
cold:
    
    hazard: Kältewelle
    
    keywords: Kältewelle, Kälteeinbruch, Kältestress, extreme Kälte, extremer Kälte, extremen Kälte, Frost, strenger Winter, Wintereinbruch, Wintereinbrüch, Kälteperiode, Kälterekord, arktische Kälte, arktischer Kälte, arktischen Kälte, Kältewarnung, Eisregen, Eiseskälte, Schneechaos
    
    hazard_event: Kältewellenereignisse
    
    definition: Eine Periode mit ungewöhnlich kaltem Wetter. In der Regel dauert eine Kältewelle zwei oder mehr Tage und kann durch starke Winde noch verstärkt werden. Die genauen Temperaturkriterien für eine Kältewelle können je nach Standort variieren.
    
heat:
    
        hazard: Hitze
        
        keywords: Hitze, extreme Temperaturen, extremen Temperaturen, Temperaturrekord, Tropennacht, Tropennächt, überhitzung, Rekordhitze, Hitzetag
        
        hazard_event: Hitzewellenereignisse
        
        definition: Eine Periode mit ungewöhnlich heißem und/oder ungewöhnlich feuchtem Wetter. In der Regel dauert eine Hitzewelle zwei oder mehr Tage. Die genauen Temperaturkriterien für eine Hitzewelle können je nach Standort variieren.
    
drought:
    
    hazard: Dürre
    
    keywords: Dürre, Rekorddürre, Trockenperiode, Trockenheit, Wasserknappheit, Niedrigwasser, Wassermangel, Niederschlagsmangel, Niederschlagsdefizit, Bodenfeuchte-Defizit, Bodenfeuchteverlust
    
    hazard_event: Dürreereignisse
    
    definition: Ein längerer Zeitraum mit ungewöhnlich geringen Niederschlägen, der zu einer Wasserknappheit für Menschen, Tiere und Pflanzen führt. Dürren unterscheiden sich von den meisten anderen Gefahren dadurch, dass sie sich langsam, manchmal sogar über Jahre hinweg, entwickeln und ihr Beginn in der Regel schwer zu erkennen ist. Dürren sind nicht nur ein physikalisches Phänomen, da ihre Auswirkungen durch menschliche Aktivitäten und den Wasserbedarf noch verstärkt werden können. Dürren werden daher oft sowohl konzeptionell als auch operativ definiert. Operative Definitionen von Dürre, d. h. der Grad der Niederschlagsverringerung, der eine Dürre ausmacht, variieren je nach Ort, Klima und Umweltbereich.
\end{lstlisting}
	\caption{Keywords and values used in the prompts for each hazard (2/3).}
	\label{fig:prompts2}
\end{figure*}
	
\begin{figure*}
\begin{lstlisting}[language=json,firstline=52,lastline=75,]
flood:
    
    hazard: Hochwasser
    
    keywords: Überschwemmung, Flut, Hochwasser, Überflutung, Flusshochwasser, Regenfälle, Sturzflut, Gletscherseeausbruch, Gletscherseeausbrüch, Gletschersee-Ausbruchsflut, Jahrhunderthochwasser
    
    hazard_event: Hochwasserereignisse
    
    definition: Ein allgemeiner Begriff für das Überlaufen von Wasser aus einem Flussbett auf normalerweise trockenes Land in der Aue (Flussüberschwemmung), überdurchschnittlich hohe Wasserstände entlang der Küste (Küstenüberschwemmung) und in Seen oder Stauseen sowie Wasseransammlungen an oder in der Nähe des Ortes, an dem der Regen gefallen ist (Sturzfluten).
    
storm:
    
    hazard: Sturm
    
    keywords: Sturm, Stürm, Unwetter, Orkan, Blizzard, Derecho, Hagel, Zyklon, Gewitter, Tornado, Mikroburst, Hurrican, Hurrikan, Taifun, Blizzard
    
    hazard_event: Sturmereignis
    
    definition: Ein Sturm ist jeder gestörte Zustand der natürlichen Umwelt oder der Atmosphäre eines astronomischen Körpers. Er kann durch erhebliche Störungen der normalen Bedingungen gekennzeichnet sein, wie z. B. starker Wind, Tornados, Hagel, Donner und Blitz (Gewitter), starke Niederschläge (Schneesturm, Regensturm), starker Eisregen (Eissturm), starke Winde (tropischer Wirbelsturm, Sturm), Wind, der bestimmte Substanzen durch die Atmosphäre transportiert, wie z. B. bei einem Staubsturm, sowie andere Formen von Unwettern.
    
\end{lstlisting}
	\caption{Keywords and values used in the prompts for each hazard (3/3).}
	\label{fig:prompts3}
\end{figure*}

\begin{figure*}
	
	\begin{lstlisting}[language=json, tabsize=2]
	Du bist ein Experte für die Klassifikation von Nachrichtenartikel bezüglich der Existenz von Referenzen auf $hazard und extreme $hazard_event.
	
	Definition von $hazard: $definition
	
	Synonyme für $hazard: $keywords.
	
	Die Nachrichtenartikel müssen mit einem dieser Labels klassifiziert werden:
	
	- Label 1: Das Dokument behandelt $hazard, extreme $hazard_event oder damit verbundene Auswirkungen.
	- Label 0: Das Dokument hat KEINE Verbindung zu $hazard oder extremen $hazard_event.
	
	Es sind nur konkrete, spezifische Naturereignisse in der realen Welt relevant. Artikel, die sich lediglich mit der Möglichkeit eines Ereignisses befassen, metaphorische Verwendungen, allgemeine Diskussionen über die Art der Gefahr oder Ereignisse sind nicht relevant.
	
	Analysiere den Inhalt des Dokuments Satz für Satz sehr sorgfaltig und vergib das Label 1 auch wenn nur ein Satz im Dokument relevant ist.
	
	Entscheide, welches Label für diesen Nachrichtenartikel das richtige ist, und beginne deine Antwort entsprechend mit 0 oder 1.
	
	Klassifiziere den folgenden Nachrichtenartikel:
	
	$text
	\end{lstlisting}
	\caption{Prompt used for the LLM experiment.}
	\label{fig:llm-prompt}
\end{figure*}

\begin{sidewaystable*}
	\centering
	\small
	\addtolength{\tabcolsep}{-0.4em}
	\begin{tabular}{llllllllllll}
		\toprule
		&  & 1 & 2 & 3 & 4 & 5 & 6 & 7 & 8 & 9 & 10 \\
		\midrule
		\multirow[t]{5}{*}{landslide} & 98 & see & bergbau & erdrutsch & wasser & rutschung & tagebau & gefahr & siedlung & erklären & bereich \\
		& 99 & tal & felssturz & fels & stein & gestein & meter & kubikmeter & berg & stürzen & groß \\
		& 80 & tote & erdrutsch & zahl & leiche & bergen & opfer & begraben & vermissen & vermisst & verschütten \\
		& 22 & mensch & erdrutsch & leben & überschwemmung & haus & schwer & behörde & heftig & sterben & region \\
		& 60 & haus & bewohner & unglück & erdrutsch & gebäude & wohnung & wohnen & früh & bürgermeisterin & ursache \\
		\cmidrule{1-12}
		\multirow[t]{2}{*}{wildfire} & 1 & feuer & brand & waldbränd & flamme & waldbrand & hektar & feuerwehr & region & groß & kontrolle \\
		& 26 & buschfeuer & australisch & koala & bundesstaat & kontinent & buschbränd & australier & tier & villa & ostküste \\
		\cmidrule{1-12}
		\multirow[t]{2}{*}{flood} & 43 & mensch & überschwemmung & haus & leben & region & dpa & behörde & stadt & land & angabe \\
		& 20 & hochwasser & sachsen & elbe & dresden & polen & pegel & meter & donau & brandenburg & tschechien \\
		\cmidrule{1-12}
		\multirow[t]{3}{*}{storm} & 16 & unwetter & wasser & hochwasser & überschwemmung & heftig & region & schwer & regen & betreffen & schaden \\
		& 13 & gewitter & wetter & regen & blitz & absagen & regnen & mark & himmel & heftig & schlecht \\
		& 7 & hurrikan & sturm & bundesstaat & treffen & kilometer & wirbelsturm & land & schaden & windgeschwindigkeit & stunde \\
		\cmidrule{1-12}
		\multirow[t]{3}{*}{drought} & 457 & dürre & notstand & ernteausfall & vieh & regenzeit & ausmaß & zentrum & versicherung & wasserreservoir & helmholtz \\
		& 246 & mitte & niedrigwasser & tag & fisch & wasser & stoff & fischsterben & sand & kreis & jugendliche \\
		& 319 & trockenheit & waldbrandgefahr & feuchtigkeit & kanton & brandgefahr & ernteausfall & nässe & stress & notstand & leiter \\
		\cmidrule{1-12}
		heat wave & 22 & grad & temperatur & hitze & celsius & tag & hitzewelle & sommer & liegen & wetter & mensch \\
		\cmidrule{1-12}
		\multirow[t]{6}{*}{cold wave} & 1 & mensch & sterben & leben & kältewelle & erfrieren & behörde & obdachlose & tote & zahl & dutzend \\
		& 194 & schneefall & heftig & stark & schneechaos & teil & fallen & sperren & schneemasse & verkehr & blockieren \\
		& 137 & eisregen & verspätung & reisend & glatt & mittag & behindern & glätte & vereist & schiene & störung \\
		& 12 & winter & kalt & mild & wehen & östlich & wintermonat & stark & atlantik & flachland & luft \\
		& 111 & kälte & warm & wärme & eisig & frieren & decke & kleidung & thermometer & anziehen & klirrend \\
		& 119 & wetter & regen & kälteeinbruch & regnen & wetterlage & kalender & bauernregel & wetterstation & eisheilig & schlecht \\
		\cline{1-12}
		\bottomrule
	\end{tabular}
	\caption{Top 10 terms of each topic considered as relevant in the \textsc{tm-b} experiments. The second column contains an arbitrary topic ID.}
	\label{tab:tm-terms}
\end{sidewaystable*}